\documentclass[conference]{IEEEtran}
\usepackage{times}

\usepackage[numbers]{natbib}
\usepackage{multicol}
\usepackage[bookmarks=true]{hyperref}
\usepackage{booktabs}
\usepackage{amsfonts}
\usepackage{nicefrac}
\usepackage{microtype}
\usepackage{xcolor}
\usepackage{amssymb}
\usepackage{multirow}
\usepackage{multicol}
\usepackage{float}
\usepackage{caption}
\usepackage{xspace}
\usepackage{amsmath}
\usepackage{graphicx}
\usepackage{wrapfig}
\usepackage{subcaption}
\usepackage{marvosym}
\usepackage{colortbl}
\usepackage{adjustbox}
\usepackage{tabularx}
\definecolor{lightblue}{RGB}{221,235,247}
\definecolor{gray94}{gray}{.94}
\definecolor{darkgreen}{RGB}{34,139,34}
\definecolor{darkred}{RGB}{160, 45, 45}
\newcommand{\green}[1]{\textcolor{darkgreen}{#1}}
\newcommand{\red}[1]{\textcolor{darkred}{#1}}
\newcommand{\gbf}[1]{\green{\bf{#1}}}
\newcommand{\rbf}[1]{\red{\bf{#1}}}
\newcommand{\grow}{\rowcolor{gray94}}
\newcommand{\brow}{\rowcolor{lightblue}}
\newcommand{\gcell}[1]{\cellcolor{gray94}{#1}} 
\newcommand{\bcell}[1]{\cellcolor{lightblue}{#1}}
\newcommand{\gray}[1]{\textcolor{gray}{#1}}
\newcommand{\ours}[0]{{H2R}\xspace}
\begin{document}
\title{H2R: A Human-to-Robot Data Augmentation for
Robot Pre-training from Videos}


\author{
\hspace*{-1.0cm}
  Guangrun Li\textsuperscript{\rm 1$^{*}$}, Yaoxu Lyu\textsuperscript{\rm 1$^{*}$}, Zhuoyang Liu\textsuperscript{\rm 1$^{*}$}, Chengkai Hou\textsuperscript{\rm 1$^{\dagger}$}, Jieyu Zhang\textsuperscript{\rm 2}, Shanghang Zhang\textsuperscript{\rm 1}~\textsuperscript{\Letter}
\vspace{0.2cm}\\
\hspace*{-1.0cm}
  \textsuperscript{\rm 1}State Key Laboratory of Multimedia Information Processing, School of Computer Science, Peking University; \\
  \hspace*{-0.5cm}
  \textsuperscript{\rm 2}University of Washington\\
  $^{*}$ Equal contribution, $^{\dagger}$ Project lead, \Letter ~Corresponding author\\
}



%

\makeatletter
\let\@oldmaketitle\@maketitle%
\renewcommand{\@maketitle}{\@oldmaketitle%
    \centering
    \setcounter{figure}{0}
    \includegraphics[width=\linewidth]{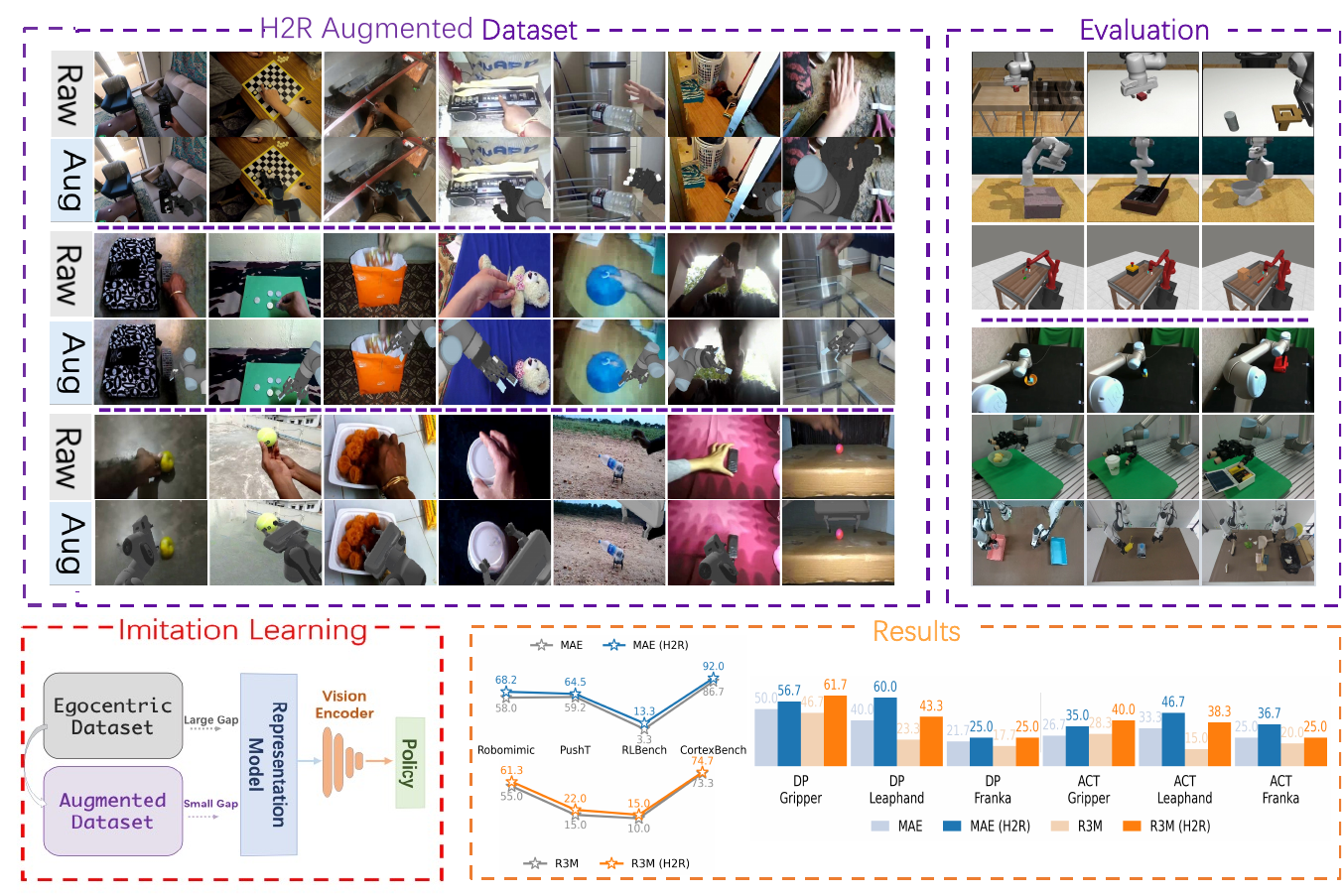}
\captionof{figure}{\small\textbf{Overview of H2R.} \textbf{H2R} is a \textbf{human-to-robot} data augmentation pipeline designed to reduce the visual gap between egocentric human-hand videos and robot-centric observations used in downstream manipulation tasks. H2R augments egocentric human videos by replacing human hands with rendered robot arms, reducing the human-to-robot visual gap during visual pre-training. The resulting encoders improve policy learning across simulation benchmarks, real-world manipulation tasks, and demonstrate consistent performance improvements.}
 \label{fig:pipeline}
}
\makeatother
\maketitle

\IEEEpeerreviewmaketitle

\begin{abstract}

Large-scale pre-training using egocentric human videos has proven effective for robot learning. However, the models pre-trained on such data can be suboptimal for robot learning due to the significant visual gap between human hands and those of different robots. To remedy this, we propose \textbf{H2R}, a human-to-robot data augmentation pipeline that converts egocentric human videos into robot-centric visual data. H2R estimates human hand pose from videos, retargets the motion to simulated robotic arms, removes human limbs via segmentation and inpainting, and composites rendered robot embodiments into the original frames with camera-aligned geometry. This process explicitly bridges the visual gap between human and robot embodiments during pre-training. 
We apply H2R to augment large-scale egocentric human video datasets such as Ego4D and SSv2. To verify the effectiveness of the augmentation pipeline, we introduce a CLIP-based image-text similarity metric that quantitatively evaluates the semantic fidelity of robot-rendered frames to the original human actions.
We evaluate H2R through comprehensive experiments in both simulation and real-world settings. In simulation, H2R consistently improves downstream success rates across four benchmark suites—Robomimic, RLBench, PushT, and CortexBench—yielding gains of 1.3\%–10.2\% across different visual encoders and policy learning methods. In real-world experiments, H2R improves performance on UR5 and dual-arm Franka/UR5 manipulation platforms, achieving 3.3\%–23.3\% success rate gains across gripper-based, dexterous, and bimanual tasks. We further demonstrate the potential of H2R in cross-embodiment generalization and its compatibility with vision–language–action models. These results indicate that H2R improves the generalization ability of robotic policies by mitigating the visual discrepancies between human and robot domains.
\end{abstract}


\section{Introduction}
\label{sec:intro}

Pre-training generalizable visual representations is a central challenge in robotic manipulation.
Recent advances in large-scale pre-training in computer vision and language
have significantly improved representation learning across domains~\cite{he2022mae,nair2022r3m,karamcheti2023voltron,datacomp,devlin2018bert,openai2024gpt4technicalreport}.
However, collecting large-scale robot demonstrations remains labor-intensive and costly~\cite{duan2023ar2,khazatsky2024droidlargescaleinthewildrobot,fang2023rh20tcomprehensiveroboticdataset},
motivating the use of readily available egocentric human videos as an alternative source for robot pre-training.

Large-scale egocentric datasets such as Ego4D~\cite{grauman2022ego4d}, Something-Something V2~\cite{goyal2017something}, and EPIC-Kitchens~\cite{damen2018scaling}
capture diverse human-object interactions and have been shown to support transferable visual representations for robotic manipulation.
However, these datasets are inherently human-centric.
The visual mismatch between human hands in egocentric videos and robotic embodiments at deployment time
introduces a gap that is not explicitly addressed during pre-training, limiting representation transfer.

To mitigate this issue, we propose \textbf{\ours} (as shown in Figure~\ref{fig:pipeline}), a simple data augmentation method that converts videos of \textbf{H}uman hand operations into that of \textbf{R}obotic arm manipulation. \ours consists of two major procedures: the first part is to generate the robotic movements to imitate the human hand movements in a video, followed by the second stage that overlays the robotic movements onto the human hand's movements in the video. Specifically, in the \texttt{first} part, we employ state-of-the-art 3D hand reconstruction model HaMeR~\cite{pavlakos2023reconstructinghands3dtransformers} to accurately detect the position and posture of the human hand in egocentric videos. Then, we simulate the same robot state in simulators to obtain the mask of robot actions. In the \texttt{second} stage, we use the Segment Anything Model~\cite{kirillov2023segment} to automatically separate human hand from background, and use the inpainting model LaMa~\cite{suvorov2021resolutionrobustlargemaskinpainting} to fill the removed hand mask. After that, we align the camera intrinsic parameters of the images detected in HaMeR with those in the simulator, and then achieve pixel-level matching between the robotic arm images in the simulators and the human hand images in the egocentric video. Finally, we overlay the robotic arm images captured by the simulator's camera onto the areas where the human hands are removed. Through such a process, \ours explicitly reduces the gap between human and robot hands by creating realistic robotic arm movements that visually mimic human hand actions. It allows the model to learn the task-specific actions demonstrated by the human hand, but with robotic arm visual representations that are more suitable for robotic systems. 

To evaluate the effectiveness of the H2R augmentation process, we introduce a CLIP-based~\cite{radford2021clip} semantic similarity metric that measures how well the rendered robot frames preserve the original action semantics. This provides a lightweight and scalable proxy to assess the alignment quality between input human videos and robot-augmented outputs. 

We further evaluate the effectiveness of \ours through a comprehensive set of experiments in both simulation and real-world settings.
In simulation, we conduct imitation learning experiments on four benchmark suites—Robomimic~\cite{mandlekar2021matters}, RLBench~\cite{james2019rlbench}, PushT~\cite{chi2023diffusionpolicy}, and CortexBench~\cite{majumdar2024searchartificialvisualcortex}—
using visual encoders pre-trained with MAE and R3M, and downstream policies trained under standard behavior cloning
and diffusion-based policy learning frameworks.
Across these benchmarks, pre-training with \ours consistently improves downstream success rates,
with average gains ranging from 1.3\% to 10.2\% depending on the encoder and task suite.

In real-world experiments, we deploy \ours-enhanced visual encoders on multiple robotic platforms,
including a UR5 arm with parallel grippers, a UR5 arm with a dexterous Leaphand, and dual-arm systems based on Franka and UR5e.
Policies are trained using both Diffusion Policy~\cite{chi2023diffusionpolicy} and ACT~\cite{zhao2023act}.
Across gripper-based, dexterous, and bimanual manipulation tasks,
\ours yields consistent performance improvements in real-world success rates,
with gains ranging from 3.3\% to 23.3\% across different encoder–policy combinations.

Beyond these primary evaluations, we conduct additional studies to characterize the generalization properties of \ours.
These include experiments on pre-training with different egocentric datasets,
cross-embodiment transfer where the augmentation embodiment differs from the downstream robot,
robustness analyses under varying demonstration densities and lighting conditions,
and integration with vision–language–action models through targeted fine-tuning.
Collectively, these results demonstrate that explicitly reducing the human-to-robot visual gap during pre-training
leads to robust and transferable improvements in downstream robotic manipulation performance.
\begin{figure*}[t]
    \centering
    \includegraphics[width=\linewidth]{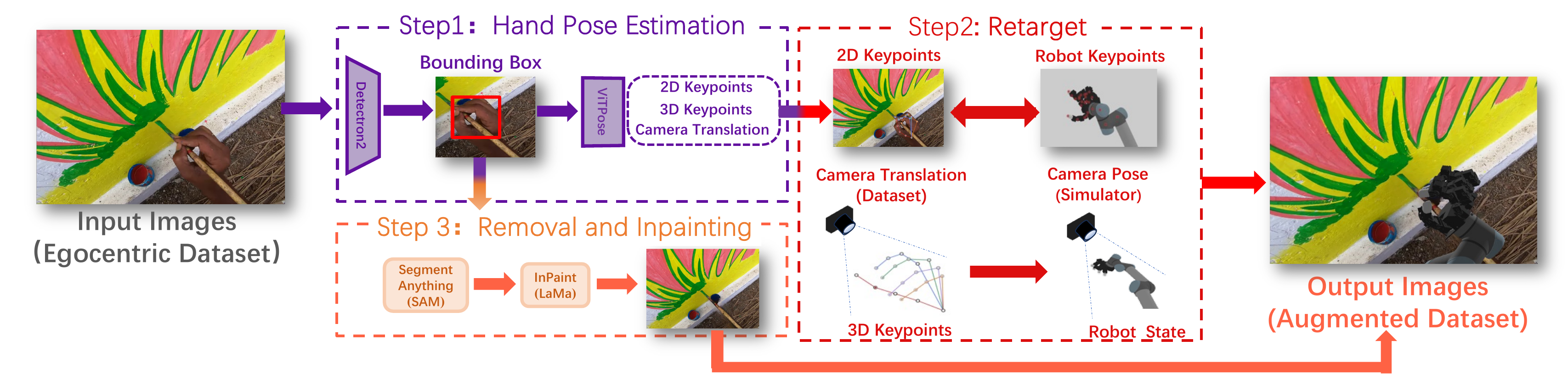}
    \vspace{-1.5em}
    \caption{\small\textbf{H2R Pipeline.} The H2R pipeline includes three steps: (1) hand pose estimation, (2) retargeting to simulate robot arm movements, and (3) removal and inpainting of the human hand to create an augmented image with the robot arm, ensuring seamless integration.}
    \vspace{-1.5em}
    \label{fig:method}
\end{figure*}

\section{Related Work}
\label{sec:relatedwork}

\noindent\textbf{Robot Imitation Learning.} Data-driven policy learning~\cite{sergey2018learning,luo2024hilserl,xiao2022mvp,yang2024equibotsim3equivariantdiffusionpolicy,chi2023diffusionpolicy,embodimentcollaboration2024openxembodimentroboticlearning} has enabled robots to autonomously perform tasks such as grasping, locomotion, and manipulation. Imitation learning~\cite{chi2023diffusionpolicy,ze2024dp3,zhao2023act,mandlekar2021matters} trains policies from successful demonstrations, often supervised by behavior cloning~\cite{torabi2018behavioral,florence2021implicit} objectives. ACT~\cite{zhao2023act} addresses non-Markovian dynamics by fusing temporal sequences, while diffusion models~\cite{chi2023diffusionpolicy,ze2024dp3} are introduced to handle the inherent multimodality of robot motions. 

\noindent\textbf{Visual Encoder Pretraining for Robotics.} Visual pretraining improves generalization of robotic policies across diverse tasks. Researchers have explored architectural designs~\cite{He_2016_CVPR,dosovitskiy12021an}, training objectives~\cite{He_2022_CVPR,chen2020simple,chen2020big}, and dataset compositions~\cite{deng2009imagenet,lin2014microsoft,shao2019object365,Schuhmann2021LAION400MOD}. PVR-Control~\cite{parisi22a} shows that pretrained visual representations can outperform direct state-based policies. RPT~\cite{radosavovic2023robot} tokenizes observations to enable masked prediction pretraining. Methods like MVP~\cite{radosavovic2022realworld} and R3M~\cite{nair2022r3m} utilize self-supervised objectives on videos to learn representations transferable to reinforcement learning. Voltron~\cite{karamcheti2023voltron} demonstrates the use of MAE and contrastive learning for hierarchical robot control.


\noindent\textbf{Cross-Domain Visual Alignment.} Bridging the domain gap between human and robot visual inputs remains a major challenge. WHIRL~\cite{bahl2022humantorobotimitationwild} matches task structure from third-person views, while RoVi-Aug~\cite{chen2024roviaugrobotviewpointaugmentation} and Mirage~\cite{chen2024mirage} manipulate appearance via segmentation or image-space preprocessing. EgoMimic~\cite{kareer2024egomimicscalingimitationlearning} removes hands and normalizes views to align egocentric perspectives. 

\section{H2R: Human-to-Robot Data Augmentation}
\label{H2R}

In this section, we present \textbf{H2R}, a human-to-robot data augmentation framework that converts egocentric human-hand videos into robot-centric visual observations.
We detail the full augmentation pipeline, including (1)hand pose estimation, (2)retargeting, and (3)removal and inpainting, and further validate the quality of the generated data using a CLIP-based semantic consistency evaluation.
An overview of the pipeline is shown in Figure~\ref{fig:method}.


\subsection{H2R Data Augmentation Pipeline}
\noindent\textbf{Step1: Hand Pose Estimation.} To overlay the human hands in the egocentric image with different robots, we first need an efficient and accurate model to detect hand information. We adopt HaMeR~\cite{pavlakos2023reconstructinghands3dtransformers}, a state-of-the-art model for 3D hand detection and reconstruction, to accurately locate the hand in the image. This step outputs the position of the human hand, keypoints, and the intrinsic and extrinsic parameters of the camera. The hand position will be used in Step 3 (Removal and Inpainting) to segment the human hand, while the keypoints and camera parameters are used in Step 2 (Retargeting) to adjust the robot's pose and camera position.

\noindent\textbf{Step2: Retargeting.} 
In this step, we simulate the robot arm to mirror the human hand's movements detected in Step 1. Using the hand keypoints and camera parameters from Step 1, we calculate the joint angles of the robot arm.

We first build the robotic arm and its end-effectors, which
are typically either grippers or dexterous hands. For dexterous hands, we estimate joint angles from hand keypoints predicted by HaMeR. Each finger joint angle is calculated using three consecutive keypoints along that finger.
For grippers, we determine how open or closed they are based on the Euclidean distance between the corresponding fingertips.
Since hand keypoints don’t capture the full arm pose (especially for joints unrelated to the hand), we manually set reasonable values for the remaining arm joints to complete a plausible robot configuration.

Next, we use the hand keypoints and camera parameters from HaMeR to adjust the camera pose in the simulator. Specifically, we define two coordinate systems: $C_H$, the coordinate system aligned with the human hand, and $C_S$, the coordinate system of the robotic arm in the simulator. By mapping the position of the camera in $C_H$ to $C_S$, we can ensure that the camera in the simulator shares the same perspective as the one captured in the real-world egocentric human image. 
The original camera position $^W\mathbf{cam}_{real}$ in the world frame is transformed to the aligned simulator position $^W\mathbf{cam}_{sim}$ using transformations from human hand ($^W_H\mathbf{R}$) and robot simulator ($^W_S\mathbf{R}$) coordinate systems:
\begin{equation}
^W\mathbf{cam}_{sim}=^W_S\mathbf{R}\;^W_H\mathbf{R}^{-1} \;^W\mathbf{cam}_{real}
\end{equation}

\noindent\textbf{Step3: Removal and Inpainting.}
After the robot pose is retargeted in Step 2, we proceed to remove the human hand from the image. We use the hand position and keypoints from Step 1 to segment out the human hand and arm regions using the Segment Anything Model (SAM)~\cite{kirillov2023segment}.

Then, to obtain clean backgrounds for inserting robotic arms, we apply LaMa~\cite{suvorov2021resolutionrobustlargemaskinpainting}, a powerful inpainting model, to fill in the removed hand-arm region. This yields clean RGB images without human limbs, providing a seamless background for inserting robotic arms in the subsequent steps.

Once both the arm segmentation from Step 2 and the inpainted image from Step 3 are available, we overlay the robot arm onto the inpainted image. We directly obtain the pixel coordinates of the human hand keypoints from Step 1. In parallel, the pixel coordinates of the robot end-effector links are computed in the simulator by projecting their 3D positions through the aligned camera using the known transformation matrices. By aligning the robot link positions with the corresponding human hand keypoints in pixel space, we ensure that the overlaid robot hand accurately matches the position and orientation of the original hand in the image, achieving precise pixel-level alignment.
\subsection{Data Quality Evaluation}
\begin{figure}[htp] 
\vspace{-0.5em}
    \centering
    \includegraphics[width=0.48\textwidth]{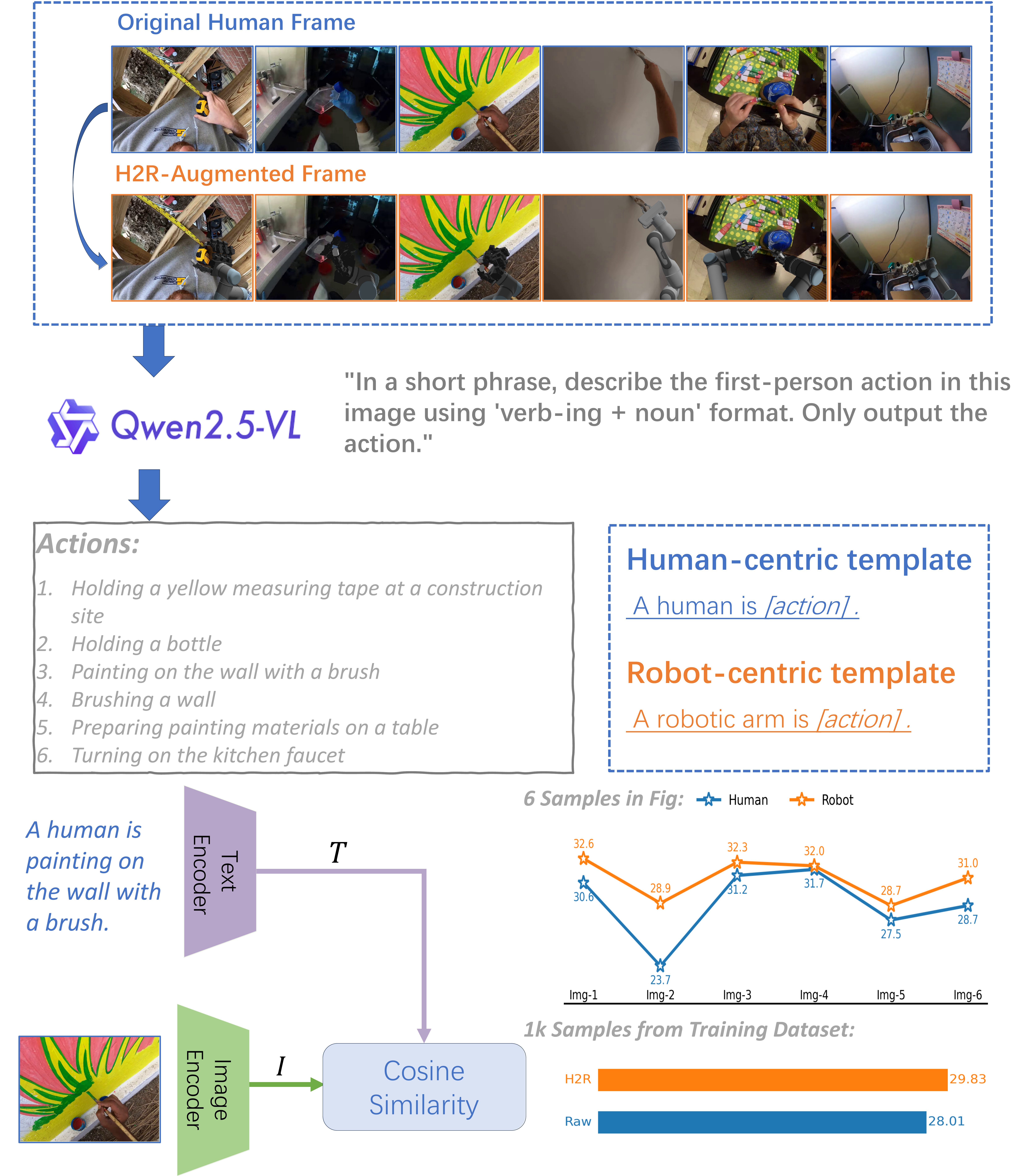}
    \caption{\small\textbf{Illustration of the CLIP-based quality evaluation.}
We evaluate augmentation quality by comparing CLIP image--text similarity under human-centric and robot-centric action prompts for original versus H2R-augmented images. The results show that H2R improves semantic alignment with robot-centric action descriptions, demonstrating the effectiveness of H2R in generating semantically coherent training data for robotic pre-training.}
\vspace{-1.5em}
    \label{fig:samples}
\end{figure}

We evaluate the visual plausibility and semantic consistency of H2R-augmented data
using a vision--language similarity metric based on CLIP~\cite{radford2021learningtransferablevisualmodels}.
The evaluation procedure is illustrated in Figure~\ref{fig:samples}, which also
shows six representative pairs of original human frames and their corresponding
H2R-augmented robot-centric images.

For each image, we associate a high-level verb--noun action description.
Specifically, given an original human frame and its H2R-augmented counterpart,
we construct two textual prompts describing the same action from different
perspectives: a human-centric prompt (\emph{``A human is [action]''}) and a
robot-centric prompt (\emph{``A robotic arm is [action]''}).
Image--text cosine similarity is then computed using CLIP ViT-B/32~\cite{radford2021learningtransferablevisualmodels}, measuring how
well each image aligns with its corresponding semantic description.

To quantify data quality at scale, we randomly sample 1,000 image pairs from the
pre-training dataset.
For each pair, the action phrase is automatically generated using
Qwen-2.5-VL~\cite{bai2025qwen25vltechnicalreport}.
We report the average CLIP similarity between original images and human-centric
prompts, and between H2R-augmented images and robot-centric prompts.

The original human images achieve an average similarity score of \textbf{28.01}, while
H2R-augmented images reach \textbf{29.83}.
This result indicates that H2R consistently improves semantic alignment with
robot-centric action descriptions.
Overall, the CLIP-based evaluation provides a scalable measure
of data quality, supporting the effectiveness of H2R in generating visually
and semantically coherent training data for robotic visual pre-training.
\section{Experiment}
 \label{sec:exp}
\newcolumntype{Y}{>{\raggedright\arraybackslash}X}
\begin{table*}[t]
\vspace{-2.5em}
\centering
\caption{\small\textbf{Summary of Experimental Settings.} This table outlines the hyperparameters, datasets, and configurations used for encoder pre-training, simulation benchmarks, and real-world robotic tasks.}
\vspace{-0.5em}
\label{table:experiment_summary}
\small

\begin{tabularx}{\textwidth}{@{}lllY@{}}
\toprule
\textbf{Phase} & \textbf{Component} & \textbf{Framework} & \textbf{Key Configurations} \\ \midrule

\multirow{4}{*}{\shortstack[l]{Encoder \\ Pretraining}}
& \multirow{2}{*}{Method}
& MAE
& ViT-B, 800 epochs, batch size 128, learning rate $4\times10^{-4}$, 8$\times$A800 GPUs. \\
& & R3M
& ViT-B, 20K steps, batch size 256, learning rate $1\times10^{-4}$, 8$\times$A800 GPUs. \\

\cmidrule(lr){2-4}

& \multirow{2}{*}{Dataset}
& SSv2
& $\sim$1M frames (subset), original vs. H2R-augmented. \\
& & Ego4D
& 117K clips subset ($\sim$1M frames), original vs. H2R-augmented. \\
\midrule

\multirow{6}{*}{\shortstack[l]{Simulation \\ Setup}}
& \multirow{4}{*}{Benchmark}
& Robomimic
& 200 epochs, 3 tasks. \\
& & RLBench
& 800 epochs, 3 tasks. \\
& & PushT
& 200 epochs, 1 task. \\
& & CortexBench
& 100 epochs, 3 tasks. \\

\cmidrule(lr){2-4}

& \multirow{2}{*}{Policy}
& BC
& 3-layer MLP (Robomimic, CortexBench), default hyperparameters. \\
& & DP
& Diffusion Policy (PushT, RLBench), default hyperparameters. \\
\midrule

\multirow{7}{*}{\shortstack[l]{Real-world \\ Setup}}
& \multirow{4}{*}{Scene}
& UR5-Gripper
& 3 tasks, keyboard-based teleoperation for collection, 1 camera, 30 demos. \\
& & UR5-Leaphand
& 3 tasks, vision-guided teleoperation for collection, 1 camera, 50 demos. \\
& & Dual-arm Franka
& 3 tasks, homogeneous-arm bilateral teleoperation for collection, 3 cameras, 300 demos. \\
& & Dual-arm UR5e
& 2 tasks, homogeneous-arm bilateral teleoperation for collection, 3 cameras, 300 demos. \\

\cmidrule(lr){2-4}

& \multirow{3}{*}{Policy}
& DP
& 300 epochs (UR5-Gripper), 3000 epochs (UR5-Leaphand), 9000 epochs (dual-arm). \\
& & ACT
& 300 epochs (UR5-Gripper), 3000 epochs (UR5-Leaphand), 9000 epochs (dual-arm). \\
& & UVA
& Default UVA configuration, 3000 epochs.\\

\bottomrule
\end{tabularx}
\vspace{-1.5em}
\end{table*}

In this section, we conduct a comprehensive experimental evaluation of H2R.
Experimental configurations are summarized in Table~\ref{table:experiment_summary}.
In simulation, we conduct imitation learning experiments across multiple benchmark suites, including studies on generalization across pre-training datasets, comparisons with robotic-data pre-training, and the effect of demonstration density.
On real robots, we perform manipulation experiments across multiple platforms and task settings, together with studies on cross-embodiment transfer, compatibility with Vision-Language-Action models, component ablations, and robustness under lighting variations.

\subsection{Experimental Setup}
\noindent\textbf{Visual Representation Pre-training.}
We adopt the MAE~\cite{he2022mae, xiao2022mvp} and R3M~\cite{nair2022r3m} frameworks
for visual encoder pre-training, both using a Vision Transformer (ViT-Base)~\cite{dosovitskiy2021imageworth16x16words}
architecture.
Pre-training is conducted on subsets of \textbf{SSv2} (1M images) or \textbf{Ego4D} (117K clips).

For each framework, we consider two settings: pre-training on the original human video data
and pre-training with additional \textbf{H2R} augmentation.
H2R overlays robot embodiments onto human videos using three representative robot types:
\textbf{UR5 with Robotiq Gripper}, \textbf{UR5 with Leaphand}, and \textbf{Franka with
 Robotiq Gripper}.
Except for cross-embodiment experiments, the robot embodiment used for H2R augmentation
is matched to the robot platform of the downstream task.

\noindent\textbf{Simulation Benchmark.}
For each pre-training method, we evaluate the performance of pre-trained encoders in imitation learning. Specifically, we select a total of 10 simulation tasks in different environments, which are from Robomimic~\cite{mandlekar2021matters}, RLBench~\cite{james2019rlbench}, PushT~\cite{chi2023diffusionpolicy} and CortexBench~\cite{majumdar2024searchartificialvisualcortex}. In particular, for Robomimic, we train the policies using the behavior cloning (BC) and evaluate them on tasks such as \textbf{MoveCan}, \textbf{Square}, and \textbf{Lift}, where the robot performs actions such as moving or lifting objects. For RLBench, we train the policies with Diffusion Policy and evaluate them on three manipulation tasks: \textbf{Close Box}, \textbf{Close Laptop Lid} and \textbf{Toilet Seat Down}. We use the \textbf{PushT} task in the Diffusion Policy evaluation framework, which evaluates a robot's ability to push an object to a target location. We also use three MetaWorld tasks: \textbf{Assembly}, \textbf{ButtonPress} and \textbf{Hammer} from VC-1 Cortex benchmark for evaluation.

\noindent\textbf{Real-world Setup.}
\begin{table*}[t]
\vspace{-2.5em}
\centering
\caption{\small\textbf{Real-world Manipulation Tasks.}
Overview of task definitions grouped by robot embodiment.}
\vspace{-0.5em}
\label{table:task_definitions}

\noindent
\begin{minipage}[t]{0.54\textwidth}
    \centering
    \vspace{0pt}
    \includegraphics[width=\linewidth]{pdf_picture/real_task.pdf}
\end{minipage}
\hfill
\begin{minipage}[t]{0.42\textwidth}
    \centering
    \vspace{0pt}
    \renewcommand{\arraystretch}{3.0}
    \begin{adjustbox}{max width=\linewidth}
\begin{tabular}{llp{4.1cm}}
\toprule
\textbf{Embodiment} & \textbf{Task Name} & \textbf{Task Description} \\
\midrule

\multirow{3}{*}{\textbf{UR5-Gripper}}
& PickCube
& The robot grasps a cube and places it into a bowl. \\

& Stack
& The robot stacks a blue cube on top of a yellow cube. \\

& CloseBox
& The robot retrieves a cube from a box, places it into a bowl, and closes the box lid. \\
\midrule

\multirow{3}{*}{\textbf{UR5-Leaphand}}
& GraspChicken
& The robot grasps a toy chicken and places it into a bowl. \\

& StandCup
& The robot grasps a fallen cup and places it upright on the table. \\

& OpenBox
& The robot opens an articulated box lid using the dexterous hand. \\
\midrule

\multirow{3}{*}{\textbf{Dual-arm Franka}}
& PlaceToy
& The left arm grasps a toy from a pink box, and the right arm transfers it into a blue box. \\

& SweepRubbish
& The left arm sweeps trash into a dustpan held by the right arm. \\

& WeighSauce
& The left arm places a cup onto a scale, while the right arm performs a pouring motion with a bottle. \\
\midrule

\multirow{2}{*}{\textbf{Dual-arm UR5e}}
& PlaceBowl
& The robot places a green bowl on top of a blue bowl. \\

& StackBlocks
& The robot stacks blocks in the order of red, yellow, and blue to form a single tower. \\
\bottomrule
\end{tabular}

    \end{adjustbox}
\end{minipage}
\vspace{-2.0em}
\end{table*}

We evaluate the effectiveness of H2R across four real-world manipulation setups:
(i) a UR5 arm equipped with a Robotiq Gripper~\cite{RobotiqAdaptiveGrippers}, 
(ii) a UR5 arm equipped with a Leaphand end effector~\cite{shaw2023leaphandlowcostefficient},
(iii) a dual-arm Franka Emika system with parallel grippers,
and (iv) a dual-arm UR5e platform configured as a human-like bimanual manipulation setup.
Across these platforms, we evaluate a diverse set of real-world manipulation tasks
spanning gripper-based, dexterous, and bimanual manipulation;
detailed task definitions are summarized in Table~\ref{table:task_definitions}.

Demonstration data are collected through human teleoperation, with the collection interface
adapted to the manipulation setting.
Specifically, gripper-based tasks are collected using keyboard-based teleoperation,
Leaphand tasks using vision-guided teleoperation following the same
setup as in CordViP~\cite{fu2025cordvip},
and dual-arm tasks using homogeneous-arm bilateral teleoperation~\cite{xu2025hactshumanascopilotteleoperationrobot}.
The number of demonstrations varies across setups and task categories,
ranging from 30 to 300 episodes depending on task complexity,
as summarized in Table~\ref{table:experiment_summary}.

For policy training, we adopt Diffusion
Policy (DP)~\cite{chi2023diffusionpolicy} and ACT~\cite{zhao2023act} as policy frameworks. We apply the pre-trained MAE and R3M visual encoders to downstream policy
learning. 

During evaluation, target objects are randomly initialized within predefined regions
following a uniform distribution consistent with expert demonstrations.
Each task is executed for 20 rollouts, and success rates are reported as the primary
evaluation metric.
\subsection{Simulation Results}
\noindent\textbf{Performance on Simulation Benchmarks.}
\begin{table*}[t]
    \centering
    \caption{\small\textbf{Simulation Benchmark Results.} Success rates (\%$\uparrow$) across diverse imitation learning suites. \textbf{Bold} indicates the best performance within each group, and \gbf{green}/\rbf{red} denotes the performance gain/drop after applying H2R.  All subsequent tables follow the same rule.}
    \vspace{-0.5em}
    \label{table:il_res}
    \setlength{\tabcolsep}{3.8pt} 
    \begin{adjustbox}{max width=\textwidth}
    \begin{tabular}{c|cccc|c|cccc|cccc}
    \toprule
    \multirow{2}{*}{\textbf{Method}} & \multicolumn{4}{c|}{\textbf{Robomimic}} & \textbf{PushT} & \multicolumn{4}{c|}{\textbf{RLBench}} & \multicolumn{4}{c}{\textbf{CortexBench}} \\ \cmidrule(lr){2-5} \cmidrule(lr){6-6} \cmidrule(lr){7-10} \cmidrule(lr){11-14}
    & \textbf{MoveCan} & \textbf{Square} & \textbf{Lift} & \textbf{Avg.} & \textbf{PushT} & \textbf{CloseBox} & \textbf{CloseLaptopLid} & \textbf{ToiletSeatDown} & \textbf{Avg.} & \textbf{Assembly} & \textbf{ButtonPress} & \textbf{Hammer} & \textbf{Avg.} \\ \midrule
    
    \grow MAE (SSv2) & 54.0 & 25.5 & 94.5 & 58.0 & 59.2 & 0.0 & 10.0 & 0.0 & 3.3 & 84.0 & 80.0 & 96.0 & 86.7 \\
    \brow \bf{MAE (H2R)} & \bf{79.5} & \bf{29.5} & \bf{95.5} & \bf{68.2} & \bf{64.5} & \bf{5.0} & \bf{15.0} & \bf{20.0} & \bf{13.3} & \bf{88.0} & \bf{88.0} & \bf{100.0} & \bf{92.0} \\
    \textit{Gain ($\Delta$)} & \gbf{+25.5} & \gbf{+4.0} & \gbf{+1.0} & \gbf{+10.2} & \gbf{+5.3} & \gbf{+5.0} & \gbf{+5.0} & \gbf{+20.0} & \gbf{+10.0} & \gbf{+4.0} & \gbf{+8.0} & \gbf{+4.0} & \gbf{+5.3} \\ \midrule

    \grow R3M (SSv2) & 59.5 & 20.5 & \bf{85.0} & 55.0 & 15.0 & 0.0 & \bf{20.0} & 10.0 & 10.0 & \bf{76.0} & 56.0 & 88.0 & 73.3 \\
    \brow \bf{R3M (H2R)} & \bf{61.5} & \bf{37.5} & \bf{85.0} & \bf{61.3} & \bf{22.0} & \bf{5.0} & \bf{20.0} & \bf{20.0} & \bf{15.0} & 68.0 & \bf{60.0} & \bf{96.0} & \bf{74.7} \\
    \textit{Gain ($\Delta$)} & \gbf{+2.0} & \gbf{+17.0} & \gray{0.0} & \gbf{+6.3} & \gbf{+7.0} & \gbf{+5.0} & \gray{0.0} & \gbf{+10.0} & \gbf{+5.0} & \rbf{-8.0} & \gbf{+4.0} & \gbf{+8.0} & \gbf{+1.3} \\ 
    \bottomrule
    \end{tabular}
    \end{adjustbox}
    \vspace{-1.5em}
\end{table*}
Table~\ref{table:il_res} shows that visual encoders pre-trained with H2R
consistently outperform those trained on the original SSv2 dataset
across all evaluated simulation benchmarks for both MAE and R3M visual encoders.
For Robomimic tasks, H2R leads to clear improvements for both encoders.
MAE achieves a 10.2\% increase in average success rate, while R3M improves by 6.3\%.
Notably, MAE observes a substantial \textbf{25.5\%} gain on the MoveCan task.
Similar trends are observed on PushT, RLBench, and CortexBench, with consistent
performance gains across the evaluated tasks for both MAE and R3M.
These results demonstrate that H2R augmentation consistently enhances the
effectiveness of visual representations learned from large-scale human video datasets
for downstream imitation learning across diverse simulation environments.

\noindent\textbf{Generalization Across Pre-training Datasets.}
\begin{table}[h]
    \centering
    \caption{\small\textbf{Simulation Results with Ego4D Pre-training.} Success rates (\%$\uparrow$) on PushT and RLBench when pre-training MAE/R3M on an Ego4D subset, comparing original vs. H2R-augmented pre-training.}
    \vspace{-0.5em}
    \label{table:il_res2}
    \setlength{\tabcolsep}{3pt} 
    \begin{adjustbox}{max width=0.48\textwidth}
    \begin{tabular}{c|c|cccc}
    \toprule
    \multirow{2}{*}{\textbf{Method}} & \textbf{PushT} & \multicolumn{4}{c}{\textbf{RLBench}} \\ \cmidrule(lr){2-2} \cmidrule(lr){3-6}
    & \textbf{PushT} & \textbf{CloseBox} & \textbf{CloseLaptopLid} & \textbf{ToiletSeatDown} & \textbf{Avg.} \\ \midrule
    
    \grow MAE (Ego4D) & 51.3 & 0.0 & 0.0 & \bf{5.0} & 1.7 \\
    \brow \bf{MAE (H2R)} & \bf{53.5} & \bf{10.0} & \bf{5.0} & 0.0 & \bf{5.0} \\
    \textit{Gain ($\Delta$)} & \gbf{+2.2} & \gbf{+10.0} & \gbf{+5.0} & \rbf{-5.0} & \gbf{+3.3} \\ \midrule

    \grow R3M (Ego4D) & \bf{13.6} & 10.0 & \bf{5.0} & 5.0 & 6.7 \\
    \brow \bf{R3M (H2R)} & \bf{13.6} & \bf{15.0} & \bf{5.0} & \bf{15.0} & \bf{11.7} \\
    \textit{Gain ($\Delta$)} & \gray{0.0} & \gbf{+5.0} & \gray{0.0} & \gbf{+10.0} & \gbf{+5.0} \\ 
    \bottomrule
    \end{tabular}
    \end{adjustbox}
    \vspace{-2.5em}
\end{table}
We evaluate H2R using Ego4D to assess its generalization beyond SSv2.
Specifically, we pre-train both MAE and R3M on an Ego4D subset of comparable scale
to SSv2, following the same training protocol as in the previous experiments.
Table~\ref{table:il_res2} reports the resulting imitation learning performance
on the PushT task and RLBench benchmarks.
On PushT, H2R leads to improvement for MAE (+2.2\%) while preserving the same
performance for R3M. On RLBench, H2R improves the average success rate for both encoders,
with MAE increasing from 1.7\% to 5.0\% and R3M from 6.7\% to 11.7\%.

These results indicate that the benefits of H2R are not specific to SSv2, and can be
consistently transferred to visual representations pre-trained on Ego4D, supporting
the general applicability of H2R across different sources of large-scale human video
data.

\noindent\textbf{Pretraining on Robotic Datasets.} We further compare H2R with visual representations pretrained directly on robotic datasets by including models pretrained on the DROID dataset~\cite{khazatsky2024droidlargescaleinthewildrobot}, following the setting reported in the recent study~\cite{jiang2024robotspretrainrobotsmanipulationcentric}.
All models use the R3M framework for pre-training, where \textbf{R3M} is pretrained on SSv2, \textbf{R3M-DROID} is pretrained on the DROID robotic dataset, and \textbf{R3M-H2R} is pretrained on the H2R-augmented SSv2 dataset.
We evaluate all representations on the same Robomimic benchmarks.
As shown in Table~\ref{table:simvsreal}, both robotic data pre-training and H2R augmentation lead to performance improvements over SSv2-only pre-training. However, R3M-H2R achieves the strongest overall results. Despite DROID being collected from real-world robotic executions, H2R pre-training
achieves stronger downstream manipulation performance, suggesting that
human-to-robot visual augmentation offers an effective alternative to direct
robotic data pretraining.
\begin{table}[h]
    \centering
    \caption{\small\textbf{Comparison to Robotic-data Pre-training.} Robomimic success rates (\%$\uparrow$) for R3M pre-training on SSv2, robotic dataset (DROID), and H2R-augmented SSv2, highlighting H2R as an alternative to direct robotic-data pre-training.}
    \vspace{-0.5em}
    \label{table:simvsreal}
    \setlength{\tabcolsep}{6pt} 
    \begin{adjustbox}{width=0.48\textwidth}
    \begin{tabular}{c|cccc}
    \toprule
    {\textbf{Method}} 
    & \textbf{MoveCan} & \textbf{Square} & \textbf{Lift} & \textbf{Avg.} \\ \midrule
    
    \grow R3M & 59.5 & 20.5 & 85.0 & 55.0 \\
    \grow R3M-DROID & 54.0 & 22.0 & \bf{96.0} & 56.7 \\
    \brow \bf{R3M-H2R} & \bf{61.5} & \bf{37.5} & 85.0 & \bf{61.3} \\
    
    \bottomrule
    \end{tabular}
    \end{adjustbox}
    \vspace{-1.5em}
\end{table}
\begin{table*}[t]
\centering
\caption{\small\textbf{Real-world Task Results.} Success rates (\%$\uparrow$) on gripper, dexterous (Leaphand), and bimanual (Franka) tasks using DP and ACT policies with MAE/R3M visual encoders. Rows denote pre-training variants,  and columns denote tasks.}
\vspace{-0.5em}
\label{table:real-world-transposed}

\begin{minipage}[t]{0.9\textwidth}
\centering
{\footnotesize{(a) DP Policy.}\par\vspace{0.2em}}
\begin{adjustbox}{max width=\textwidth}
\begin{tabular}{c|c|cccc|cccc|cccc}
\toprule
\multirow{3}{*}{\bf{Policy}} & \multirow{3}{*}{\bf{Method}}
& \multicolumn{12}{c}{\bf{Tasks}} \\
\cmidrule(lr){3-14}
& & \multicolumn{4}{c|}{\bf{Gripper}}
  & \multicolumn{4}{c|}{\bf{Leaphand}}
  & \multicolumn{4}{c}{\bf{Franka}} \\
\cmidrule(lr){3-6}\cmidrule(lr){7-10}\cmidrule(lr){11-14}
& & \bf{PickCube} & \bf{Stack} & \bf{CloseBox} & \bf{Avg.}
  & \bf{GraspChicken} & \bf{StandCup} & \bf{OpenBox} & \bf{Avg.}
  & \bf{PlaceToy} & \bf{SweepRubbish} & \bf{WeighSauce} & \bf{Avg.} \\
\midrule

\multirow{6}{*}{\bf{DP}}
& \gcell{MAE (SSv2)}
& \gcell{45.0} & \gcell{50.0} & \gcell{\bf{55.0}} & \gcell{50.0}
& \gcell{40.0} & \gcell{35.0} & \gcell{45.0} & \gcell{40.0}
& \gcell{10.0} & \gcell{\bf{15.0}} & \gcell{40.0} & \gcell{21.7}\\

& \bcell{\bf{MAE (H2R)}}
& \bcell{\bf{65.0}} & \bcell{\bf{55.0}} & \bcell{50.0} & \bcell{\bf{56.7}}
& \bcell{\bf{55.0}} & \bcell{\bf{60.0}} & \bcell{\bf{65.0}} & \bcell{\bf{60.0}}
& \bcell{\bf{15.0}} & \bcell{\bf{15.0}} & \bcell{\bf{45.0}} & \bcell{\bf{25.0}}\\

& \textit{Gain ($\Delta$)}
& \gbf{+20.0} & \gbf{+5.0} & \rbf{-5.0} & \gbf{+6.7}
& \gbf{+15.0} & \gbf{+25.0} & \gbf{+20.0} & \gbf{+20.0}
& \gbf{+5.0} & \gray{0.0} & \gbf{+5.0} & \gbf{+3.3}\\

\cmidrule(lr){2-14}

& \gcell{R3M (SSv2)}
& \gcell{40.0} & \gcell{55.0} & \gcell{45.0} & \gcell{46.7}
& \gcell{10.0} & \gcell{20.0} & \gcell{40.0} & \gcell{23.3}
& \gcell{10.0} & \gcell{\bf{20.0}} & \gcell{20.0} & \gcell{17.7}\\

& \bcell{\bf{R3M (H2R)}}
& \bcell{\bf{50.0}} & \bcell{\bf{70.0}} & \bcell{\bf{65.0}} & \bcell{\bf{61.7}}
& \bcell{\bf{35.0}} & \bcell{\bf{50.0}} & \bcell{\bf{45.0}} & \bcell{\bf{43.3}}
& \bcell{\bf{25.0}} & \bcell{\bf{20.0}} & \bcell{\bf{30.0}} & \bcell{\bf{25.0}}\\

& \textit{Gain ($\Delta$)}
& \gbf{+10.0} & \gbf{+15.0} & \gbf{+20.0} & \gbf{+15.0}
& \gbf{+25.0} & \gbf{+30.0} & \gbf{+5.0} & \gbf{+20.0}
& \gbf{+15.0} & \gray{0.0} & \gbf{+10.0} & \gbf{+7.3} \\

\bottomrule
\end{tabular}
\end{adjustbox}
\end{minipage}

\vspace{0.8em}

\begin{minipage}[t]{0.9\textwidth}
\centering
{\footnotesize(b) ACT Policy.\par\vspace{0.2em}}
\begin{adjustbox}{max width=\textwidth}
\begin{tabular}{c|c|cccc|cccc|cccc}
\toprule
\multirow{3}{*}{\bf{Policy}} & \multirow{3}{*}{\bf{Method}}
& \multicolumn{12}{c}{\bf{Tasks}} \\
\cmidrule(lr){3-14}
& & \multicolumn{4}{c|}{\bf{Gripper}}
  & \multicolumn{4}{c|}{\bf{Leaphand}}
  & \multicolumn{4}{c}{\bf{Franka}} \\
\cmidrule(lr){3-6}\cmidrule(lr){7-10}\cmidrule(lr){11-14}
& & \bf{PickCube} & \bf{Stack} & \bf{CloseBox} & \bf{Avg.}
  & \bf{GraspChicken} & \bf{StandCup} & \bf{OpenBox} & \bf{Avg.}
  & \bf{PlaceToy} & \bf{SweepRubbish} & \bf{WeighSauce} & \bf{Avg.} \\
\midrule

\multirow{6}{*}{\bf{ACT}}
& \gcell{MAE (SSv2)}
& \gcell{25.0} & \gcell{20.0} & \gcell{35.0} & \gcell{26.7}
& \gcell{45.0} & \gcell{25.0} & \gcell{30.0} & \gcell{33.3}
& \gcell{25.0} & \gcell{20.0} & \gcell{30.0} & \gcell{25.0} \\

& \bcell{\bf{MAE (H2R)}}
& \bcell{\bf{30.0}} & \bcell{\bf{35.0}} & \bcell{\bf{40.0}} & \bcell{\bf{35.0}}
& \bcell{\bf{50.0}} & \bcell{\bf{50.0}} & \bcell{\bf{40.0}} & \bcell{\bf{46.7}}
& \bcell{\bf{40.0}} & \bcell{\bf{35.0}} & \bcell{\bf{35.0}} & \bcell{\bf{36.7}} \\

& \textit{Gain ($\Delta$)}
& \gbf{+5.0} & \gbf{+15.0} & \gbf{+5.0} & \gbf{+8.3}
& \gbf{+5.0} & \gbf{+25.0} & \gbf{+10.0} & \gbf{+13.3}
& \gbf{+15.0} & \gbf{+15.0} & \gbf{+5.0} & \gbf{+11.7} \\

\cmidrule(lr){2-14}

& \gcell{R3M(SSv2)}
& \gcell{25.0} & \gcell{20.0} & \gcell{40.0} & \gcell{28.3}
& \gcell{10.0} & \gcell{20.0} & \gcell{15.0} & \gcell{15.0}
& \gcell{20.0} & \gcell{\bf{20.0}} & \gcell{20.0} & \gcell{20.0} \\

& \bcell{\bf{R3M (H2R)}}
& \bcell{\bf{30.0}} & \bcell{\bf{40.0}} & \bcell{\bf{50.0}} & \bcell{\bf{40.0}}
& \bcell{\bf{35.0}} & \bcell{\bf{60.0}} & \bcell{\bf{20.0}} & \bcell{\bf{38.3}}
& \bcell{\bf{30.0}} & \bcell{\bf{20.0}} & \bcell{\bf{25.0}} & \bcell{\bf{25.0}} \\

& \textit{Gain ($\Delta$)}
& \gbf{+5.0} & \gbf{+20.0} & \gbf{+10.0} & \gbf{+11.7}
& \gbf{+25.0} & \gbf{+40.0} & \gbf{+5.0} & \gbf{+23.3}
& \gbf{+10.0} & \gray{0.0} & \gbf{+5.0} & \gbf{+5.0} \\

\bottomrule
\end{tabular}
\end{adjustbox}
\end{minipage}
\vspace{-1.5em}
\end{table*}

\begin{table}[h]
    \centering
    \caption{\small\textbf{Effects of Demonstration Density.} Success rates (\%$\uparrow$) under sparse vs. dense RLBench demonstrations, comparing MAE encoders pre-trained with and without H2R.}
    \vspace{-0.5em}
    \label{table:density}
    \setlength{\tabcolsep}{2.5pt} 
    \begin{adjustbox}{max width=0.48\textwidth}
    \begin{tabular}{c|c|cccc}
        \toprule
        \textbf{Data} & \textbf{Model} & \textbf{CloseBox} & \textbf{CloseLaptopLid} & \textbf{ToiletSeatDown} & \textbf{Avg.} \\ \midrule
        \multirow{3}{*}{Sparse data}
        & \gcell{MAE} & \gcell{0.0} & \gcell{10.0} & \gcell{0.0} & \gcell{3.3} \\
        & \bcell{\textbf{MAE (H2R)}} & \bcell{\textbf{5.0}} & \bcell{\textbf{15.0}} & \bcell{\textbf{20.0}} & \bcell{\textbf{13.3}} \\
        & \textit{Gain ($\Delta$)} & \gbf{+5.0} & \gbf{+5.0} & \gbf{+20.0} & \gbf{+10.0} \\ \midrule
        \multirow{3}{*}{Dense data}
        & \gcell{MAE} & \gcell{50.0} & \gcell{40.0} & \gcell{45.0} & \gcell{45.0} \\
        & \bcell{\textbf{MAE (H2R)}} & \bcell{\textbf{60.0}} & \bcell{\textbf{45.0}} & \bcell{\textbf{60.0}} & \bcell{\textbf{55.0}} \\
        & \textit{Gain ($\Delta$)} & \gbf{+10.0} & \gbf{+5.0} & \gbf{+15.0} & \gbf{+10.0} \\ 
        \bottomrule
    \end{tabular}
    \end{adjustbox}
    \vspace{-2.5em}
\end{table}

\noindent\textbf{Effects of Demonstration Density.}
We observe relatively low success rates on RLBench under sparse, keypoint-based supervision, a setting commonly adopted in prior RLBench works to improve training efficiency~\cite{goyal2023rvt, jia2024lift3dfoundationpolicylifting, shridhar2023peract}.
This observation motivates an additional study to examine whether the effectiveness of H2R depends on demonstration density, by comparing sparse and dense RLBench training under identical settings.

As shown in Table~\ref{table:density}, increasing demonstration density substantially improves overall task performance (e.g., MAE average success rate increases from 3.3\% to 45.0\%). More importantly, H2R consistently yields performance gains in both sparse and dense regimes. This result indicates that H2R is not tailored to a specific data density, but instead provides robust visual representations that remain effective across different demonstration settings.
\subsection{Real-world Results}
\noindent\textbf{Performance on Real-world Manipulation Tasks.}
In real-world experiments, we evaluate H2R on three categories of manipulation tasks-gripper, dexterous, and bimanual—using corresponding robotic platforms, including a UR5 equipped with a Robotiq gripper, a UR5 equipped with a Leaphand end effector, and a dual-arm Franka system.
We adopt Diffusion Policy (DP)~\cite{chi2023diffusionpolicy} and ACT~\cite{zhao2023act} as policy frameworks, with visual encoders pre-trained using MAE and R3M.
In this setting, the robot embodiment used for H2R augmentation during pre-training is identical to that used in downstream policy training and evaluation.

As shown in Table~\ref{table:real-world-transposed}, H2R consistently improves real-world success rates across all task categories, encoders, and policies.
Notably, the most pronounced gains are observed on Leaphand tasks, where H2R improves average performance by \textbf{20.0\%} (MAE, R3M) under DP, and by \textbf{13.3\%} (MAE) and \textbf{23.3\%} (R3M) under ACT.
Gripper-based and dual-arm Franka tasks also benefit from H2R, with consistent improvements across both MAE and R3M.
These results demonstrate that H2R effectively enhances real-world manipulation performance.

\begin{table}[h]
\centering
\caption{\small\textbf{Real-world results with Ego4D pre-training.} Success rates (\%$\uparrow$) on UR5-Leaphand tasks using ACT, comparing Ego4D pre-training with vs. without H2R augmentation.}
\vspace{-0.5em}
\label{table:real-ego4d}
\setlength{\tabcolsep}{5.0pt}
\begin{adjustbox}{width=0.48\textwidth}
\begin{tabular}{c|cccc}
\toprule
\bf{Method} & \bf{GraspChicken} & \bf{StandCup} & \bf{OpenBox} & \bf{Avg.} \\
\midrule

\gcell{MAE (Ego4D)}
& \gcell{25.0} & \gcell{25.0} & \gcell{35.0} & \gcell{28.3} \\
\bcell{\bf{MAE (H2R)}}
& \bcell{\bf{35.0}} & \bcell{\bf{50.0}} & \bcell{\bf{45.0}} & \bcell{\bf{43.3}} \\
\textit{Gain ($\Delta$)}
& \gbf{+10.0} & \gbf{+25.0} & \gbf{+10.0} & \gbf{+15.0} \\

\cmidrule(lr){1-5}

\gcell{R3M (Ego4D)}
& \gcell{15.0} & \gcell{20.0} & \gcell{30.0} & \gcell{21.7} \\
\bcell{\bf{R3M (H2R)}}
& \bcell{\bf{20.0}} & \bcell{\bf{25.0}} & \bcell{\bf{40.0}} & \bcell{\bf{28.3}} \\
\textit{Gain ($\Delta$)}
& \gbf{+5.0} & \gbf{+5.0} & \gbf{+10.0} & \gbf{+6.7} \\

\bottomrule
\end{tabular}
\end{adjustbox}
\vspace{-1.5em}
\end{table}

\begin{table*}[h]
\vspace{-1.0em}
\centering
\caption{\small\textbf{Cross-Embodiment Real-world Results.}
UR5-Leaphand task success rates (\%$\uparrow$) when the robot embodiment used for H2R augmentation during pre-training differs from the downstream embodiment (UR5-based vs. Franka-based augmentation).}
\vspace{-0.5em}
\label{table:cross-embodiment-transposed}

\begin{minipage}[t]{0.49\textwidth}
\centering
{\footnotesize{(a) DP Policy.}\par\vspace{0.2em}}
\begin{adjustbox}{max width=\textwidth}
\begin{tabular}{c|c|cccc}
\toprule
\bf{Policy} & \bf{Method}
& \bf{GraspChicken} & \bf{StandCup} & \bf{OpenBox} & \bf{Avg.} \\
\midrule

\multirow{10}{*}{\bf{DP}}
& \gcell{MAE (SSv2)}
& \gcell{40.0} & \gcell{35.0} & \gcell{45.0} & \gcell{40.0} \\

& \bcell{\bf{MAE (H2R-UR5)}}
& \bcell{\bf{55.0}} & \bcell{\bf{60.0}} & \bcell{\bf{65.0}} & \bcell{\bf{60.0}} \\

& \textit{Gain ($\Delta$)}
& \gbf{+15.0} & \gbf{+25.0} & \gbf{+20.0} & \gbf{+20.0} \\

& \bcell{MAE (H2R-Franka)}
& \bcell{35.0} & \bcell{50.0} & \bcell{45.0} & \bcell{43.3} \\

& \textit{Gain ($\Delta$)}
& \rbf{-5.0} & \gbf{+15.0} & \gray{0.0} & \gbf{+3.3} \\

\cmidrule(lr){2-6}

& \gcell{R3M (SSv2)}
& \gcell{10.0} & \gcell{20.0} & \gcell{40.0} & \gcell{23.3} \\

& \bcell{\bf{R3M(H2R-UR5)}}
& \bcell{\bf{35.0}} & \bcell{\bf{50.0}} & \bcell{\bf{45.0}} & \bcell{\bf{43.3}} \\

& \textit{Gain ($\Delta$)}
& \gbf{+25.0} & \gbf{+30.0} & \gbf{+5.0} & \gbf{+20.0} \\

& \bcell{R3M(H2R-Franka)}
& \bcell{20.0} & \bcell{30.0} & \bcell{45.0} & \bcell{31.7} \\

& \textit{Gain ($\Delta$)}
& \gbf{+10.0} & \gbf{+10.0} & \gbf{+5.0} & \gbf{+10.0} \\

\bottomrule
\end{tabular}
\end{adjustbox}
\end{minipage}
\hfill
\begin{minipage}[t]{0.49\textwidth}
\centering
{\footnotesize{(b) ACT Policy.}\par\vspace{0.2em}}
\begin{adjustbox}{max width=\textwidth}
\begin{tabular}{c|c|cccc}
\toprule
\bf{Policy} & \bf{Method}
& \bf{GraspChicken} & \bf{StandCup} & \bf{OpenBox} & \bf{Avg.} \\
\midrule

\multirow{10}{*}{\bf{ACT}}
& \gcell{MAE (SSv2)}
& \gcell{45.0} & \gcell{25.0} & \gcell{30.0} & \gcell{33.3} \\

& \bcell{\bf{MAE (H2R-UR5)}}
& \bcell{\bf{50.0}} & \bcell{\bf{50.0}} & \bcell{\bf{40.0}} & \bcell{\bf{46.7}} \\

& \textit{Gain ($\Delta$)}
& \gbf{+5.0} & \gbf{+25.0} & \gbf{+10.0} & \gbf{+13.3} \\

& \bcell{MAE (H2R-Franka)}
& \bcell{\bf{50.0}} & \bcell{\bf{50.0}} & \bcell{30.0} & \bcell{43.3} \\

& \textit{Gain ($\Delta$)}
& \gbf{+5.0} & \gbf{+25.0} & \gray{0.0} & \gbf{+10.0} \\

\cmidrule(lr){2-6}

& \gcell{R3M (SSv2)}
& \gcell{10.0} & \gcell{20.0} & \gcell{15.0} & \gcell{15.0} \\

& \bcell{\bf{R3M (H2R-UR5)}}
& \bcell{35.0} & \bcell{\bf{60.0}} & \bcell{\bf{20.0}} & \bcell{\bf{38.3}} \\

& \textit{Gain ($\Delta$)}
& \gbf{+25.0} & \gbf{+40.0} & \gbf{+5.0} & \gbf{+23.3} \\

& \bcell{R3M (H2R-Franka)}
& \bcell{\bf{40.0}} & \bcell{25.0} & \bcell{5.0} & \bcell{23.3} \\

& \textit{Gain ($\Delta$)}
& \gbf{+30.0} & \gbf{+5.0} & \rbf{-10.0} & \gbf{+8.3} \\

\bottomrule
\end{tabular}
\end{adjustbox}
\end{minipage}
\vspace{-1.5em}
\end{table*}

\noindent\textbf{Generalization Across Pre-training Datasets.}
To examine whether the effectiveness of H2R observed in simulation also extends to
real-world settings, we further evaluate visual encoders pre-trained on Ego4D in
real-world Leaphand manipulation tasks.
Following the same pre-training protocol as in the simulation experiments, both
MAE and R3M are trained on an Ego4D subset of comparable scale, with and without
H2R augmentation, and deployed using the ACT policy.

Table~\ref{table:real-ego4d} reports the success rates across three real-world
Leaphand tasks.
For MAE, applying H2R consistently improves performance on all tasks, increasing
the average success rate from 28.3\% to 43.3\%.
For R3M, H2R also yields consistent gains, with the average success rate improving
from 21.7\% to 28.3\%.

These results are consistent with the trends observed in simulation and indicate
that the benefits of H2R extend to real-world manipulation when visual encoders are
pre-trained on Ego4D.

\noindent\textbf{Cross-embodiment Generalization.}
We further evaluate the cross-embodiment generalization of H2R by applying
H2R-augmented pre-training using a robot embodiment that differs from the one used
in downstream policy training and evaluation.
Specifically, we consider two augmentation settings during pre-training:
UR5-based H2R and Franka-based H2R, while all downstream policies are trained and
evaluated on UR5-Leaphand tasks.

As shown in Table~\ref{table:cross-embodiment-transposed}, H2R continues to improve
performance under embodiment mismatch for both MAE and R3M encoders, as well as
for both DP and ACT policies.
Using UR5-based H2R augmentation yields the strongest gains, with average success
rates increasing by \textbf{20.0\%} (MAE) and \textbf{21.7\%} (R3M) under DP, and by
\textbf{13.3\%} (MAE) and \textbf{23.3\%} (R3M) under ACT.
When H2R is applied using Franka-based augmentation, performance improvements
remain consistent but are generally smaller in magnitude.

These results suggest that the effectiveness of H2R is not strictly limited
to cases where the pre-training and downstream embodiments are identical.
Within the evaluated settings, H2R remains beneficial even when the augmentation
embodiment differs from that used in policy learning, indicating a certain degree of
robustness to embodiment variation.

\noindent\textbf{Synergy with Vision-Language-Action Models.}
A common paradigm in Vision-Language-Action (VLA) models is to initialize the visual backbone using encoders pre-trained on large-scale datasets (e.g., CLIP, VAEs). In this experiment, we investigate whether fine-tuning a pre-trained visual backbone on robot-oriented video data can enhance its suitability for robotic policy learning. Specifically, we fine-tune the pre-trained VAE encoder and decoder used in the Unified Video Action Model(UVA)~\cite{li2025unifiedvideoactionmodel} separately on both the original SSv2 and the H2R-augmented datasets. The fine-tuned VAEs are subsequently frozen and integrated into the UVA framework for policy training on two distinct dual-arm UR5e robot tasks. The results shown in Table ~\ref{tab:uva_backbone_ablation} demonstrate a clear performance trend in the different variants of the UVA model.
\begin{table}[h]
\centering
\caption{\small\textbf{UVA  with different visual backbones.} Success rates (\%$\uparrow$) on dual-arm UR5e tasks when using UVA with original backbone, and with VAEs fine-tuned on SSv2 or H2R-augmented data.}
\vspace{-0.5em}
\label{tab:uva_backbone_ablation}
\setlength{\tabcolsep}{6pt}
\begin{adjustbox}{width=0.38\textwidth}
\begin{tabular}{c|ccc}
\toprule
\bf{Method} & \bf{PlaceBowl} & \bf{StackBlocks} & \bf{Avg.} \\
\midrule

\grow UVA
& 0.20 & 0.20 & 0.20 \\

\grow UVA (SSv2)
& 0.10 & 0.20 & 0.15 \\

\brow \bf{UVA (H2R)}
& \bf{0.40} & \bf{0.30} & \bf{0.35} \\

\bottomrule
\end{tabular}
\end{adjustbox}
\vspace{-1.5em}
\end{table}

\begin{table}[h]
\centering
\caption{\small\textbf{Ablation Study.}
Ablation results by removing robot overlay (\textbf{w/o Overlay}) and camera-hand retargeting (\textbf{w/o Retarget}).}
\vspace{-0.5em}
\label{table:Ablation}
\setlength{\tabcolsep}{5.0pt}
\begin{adjustbox}{width=0.48\textwidth}
\begin{tabular}{c|c|cccc}
\toprule
\bf{Policy} & \bf{Method} & \bf{GraspChicken} & \bf{StandCup} & \bf{OpenBox} & \bf{Avg.} \\
\midrule

\multirow{5}{*}{\bf{DP}}
& \bcell{\bf{H2R}}
& \bcell{\bf{55.0}} & \bcell{\bf{60.0}} & \bcell{\bf{65.0}} & \bcell{\bf{60.0}} \\

& \gcell{H2R w/o Overlay}
& \gcell{30.0} & \gcell{40.0} & \gcell{20.0} & \gcell{30.0} \\

& \textit{Gain ($\Delta$)}
& \rbf{-25.0} & \rbf{-20.0} & \rbf{-45.0} & \rbf{-30.0} \\

& \gcell{H2R w/o Retarget}
& \gcell{30.0} & \gcell{55.0} & \gcell{45.0} & \gcell{43.3} \\

& \textit{Gain ($\Delta$)}
& \rbf{-25.0} & \rbf{-5.0} & \rbf{-20.0} & \rbf{-16.7} \\

\midrule

\multirow{5}{*}{\bf{ACT}}
& \bcell{\bf{H2R}}
& \bcell{\bf{50.0}} & \bcell{\bf{50.0}} & \bcell{\bf{40.0}} & \bcell{\bf{46.7}} \\

& \gcell{H2R w/o Overlay}
& \gcell{25.0} & \gcell{35.0} & \gcell{25.0} & \gcell{28.3} \\

& \textit{Gain ($\Delta$)}
& \rbf{-25.0} & \rbf{-15.0} & \rbf{-15.0} & \rbf{-18.3} \\

& \gcell{H2R w/o Retarget}
& \gcell{45.0} & \gcell{30.0} & \gcell{15.0} & \gcell{30.0} \\

& \textit{Gain ($\Delta$)}
& \rbf{-5.0} & \rbf{-20.0} & \rbf{-25.0} & \rbf{-16.7} \\

\bottomrule
\end{tabular}
\end{adjustbox}
\vspace{-1.0em}
\end{table}

\noindent\textbf{Ablation Study.}
To evaluate the effectiveness of each component in H2R, we conduct ablation studies on two time-consuming steps: (1) performing hand inpainting without overlaying a robotic arm (H2R  w/o Overlay), and (2) overlaying the arm without precise alignment between the hand and the camera, instead using random pasting (H2R w/o Retarget). Table~\ref{table:Ablation} shows the necessity and effectiveness of each component in H2R.
The first step leads to a significant drop in success rate due to the loss of critical human-object interaction pixels after inpainting. The second step fails to provide accurate motion cues for the model and introduces visual mismatches with real-world manipulation tasks. 

\begin{table}[t]
\centering
\caption{\small\textbf{Generalization under Lighting Variations.}
Success rates (\%$\uparrow$) on UR5-Leaphand tasks under real-world lighting disturbances, comparing no augmentation, H2R, and H2R with lighting augmentation (H2R+LightAug).}
\vspace{-0.5em}
\label{table:light}
\setlength{\tabcolsep}{5.0pt}
\begin{adjustbox}{width=0.48\textwidth}
\begin{tabular}{c|cccc}
\toprule
\bf{Method} & \bf{GraspChicken} & \bf{StandCup} & \bf{OpenBox} & \bf{Avg.} \\
\midrule

\gcell{MAE}
& \gcell{10.0} & \gcell{20.0} & \gcell{5.0} & \gcell{11.7} \\

\bcell{MAE (H2R)}
& \bcell{10.0} & \bcell{25.0} & \bcell{0.0} & \bcell{11.7} \\

\textit{Gain ($\Delta$)}
& \gray{0.0} & \gbf{+5.0} & \rbf{-5.0} & \gray{0.0} \\

\bcell{\bf{MAE (H2R + LightAug)}}
& \bcell{\bf{20.0}} & \bcell{\bf{40.0}} & \bcell{\bf{25.0}} & \bcell{\bf{28.3}} \\

\textit{Gain ($\Delta$)}
& \gbf{+10.0} & \gbf{+20.0} & \gbf{+20.0} & \gbf{+16.6} \\

\cmidrule(lr){1-5}

\gcell{R3M}
& \gcell{0.0} & \gcell{5.0} & \gcell{10.0} & \gcell{5.0} \\

\bcell{R3M (H2R)}
& \bcell{0.0} & \bcell{15.0} & \bcell{\bf{10.0}} & \bcell{8.3} \\

\textit{Gain ($\Delta$)}
& \gray{0.0} & \gbf{+10.0} & \gray{0.0} & \gbf{+3.3} \\

\bcell{\bf{R3M (H2R + LightAug)}}
& \bcell{\bf{10.0}} & \bcell{\bf{20.0}} & \bcell{\bf{10.0}} & \bcell{\bf{13.3}} \\

\textit{Gain ($\Delta$)}
& \gbf{+10.0} & \gbf{+15.0} & \gray{0.0} & \gbf{+8.3} \\

\bottomrule
\end{tabular}
\end{adjustbox}
\vspace{-2.5em}
\end{table}

\noindent\textbf{Robustness to Lighting Variations.}
To evaluate the generalization under varying lighting conditions, we introduce illumination disturbances during evaluation. Additionally, during training, we incorporate randomized lighting with varying directions and colors into the simulation environment for data augmentation. We compare three settings: no augmentation(MAE, R3M), H2R augmentation(H2R), and H2R with lighting disturbances(H2R+LightAug). As shown in Table~\ref{table:light}, the model trained with H2R and lighting perturbations demonstrates significantly better generalization to real-world lighting variations than other baselines, highlighting the effectiveness of H2R in bridging the domain gap caused by lighting variations.

\section{Conclusion}
\label{sec:conclusion}
We propose H2R, a data augmentation technique that bridges the visual gap between human hand demonstrations and robotic arm manipulations by replacing human hands in first-person videos with robotic arm movements. Using 3D hand reconstruction and image inpainting models, H2R generates synthetic robotic arm manipulation sequences, making them more suitable for robot pre-training. Experiments across simulation benchmarks and real-world tasks demonstrate consistent improvements in success rates for encoders trained with various pre-training methods (e.g., MAE, R3M), highlighting the effectiveness and generalization of H2R. H2R enables efficient transfer of task knowledge from human demonstrations to robotic systems, reducing the reliance on costly robot-specific data collection. 


\bibliographystyle{plainnat}
\bibliography{references}

\newpage
\appendix
\setcounter{page}{1}
\begin{figure*}[h] 
    \centering
    \includegraphics[width=\textwidth]{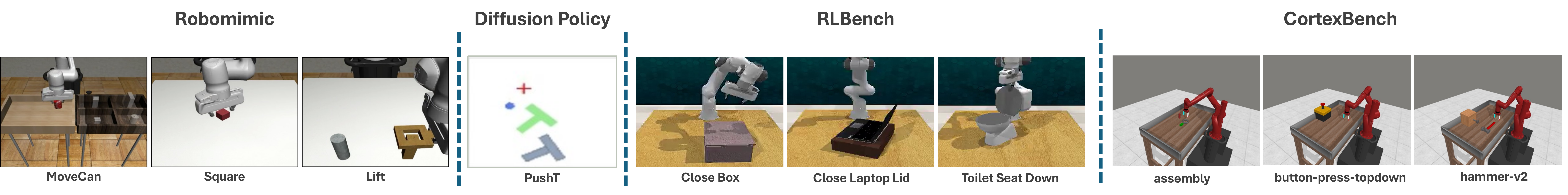}
    \caption{\textbf{Simulation benchmark.} We choose 3 tasks from the Robomimic, 3 tasks from the RLBench, and 3 tasks from the CortexBench, covering a range of robotic manipulation skills. We also include the PushT task, designed for the Diffusion Policy framework, as an additional benchmark to evaluate performance in a different task setup.} 
    \label{fig:simulator}
\end{figure*}
\subsection{Details of Simulator Camera Position Alignment}
\label{sec:camera}
We define two coordinate systems: $C_H$, the coordinate system aligned with the human hand, and $C_S$, the coordinate system of the robot arm in the simulator. We build the coordinate system $^W\mathbf{I}_H$ based on the hand keypoints: 
\begin{equation}
    ^{W}\mathbf{I}_H=\{^w\mathbf{i}_{H,x},^w\mathbf{i}_{H,y},^w\mathbf{i}_{H,z} \}
\end{equation}

Where $^w\mathbf{i}_{H,x},^w\mathbf{i}_{H,y},^w\mathbf{i}_{H,z}$ are unit vectors along the x-axes, y-axes and z-axes of the human hand coorinate system. With the keypoints get in HaMeR, we build the three axis of coordinates with the following functions:
\begin{equation}
\begin{aligned}
    ^{w}\mathbf{i}_{H,x} & =^{w}\mathbf{i}_{0,9} \\
    ^{w}\mathbf{i}_{H,y} & =^{w}\mathbf{i}_{0,9}\times ^{w}\mathbf{i}_{0,13} \\
    ^{w}\mathbf{i}_{H,z} & =^{w}\mathbf{i}_{H,x} \times ^{w}\mathbf{i}_{H,y}
\end{aligned}
\end{equation}

Where $^{w}\mathbf{i}_{0,9}$ and $^{w}\mathbf{i}_{0,13}$ are unit vectors along the middle and ring fingers, respectively. In this notation, the first index (0) refers to the specific finger (middle or ring), and the second index (9 and 13) corresponds to the joint numbers along those fingers, as defined by the MANO model. Similarly, To construct the mapping from hand pose to robot arms, we need to get another coordinate system $^W\mathbf{I}_S$ in the simulator: 
\begin{equation}
    ^{W}\mathbf{I}_S=\{^w\mathbf{i}_{S,x},^w\mathbf{i}_{S,y},^w\mathbf{i}_{S,z} \}
\end{equation}
The method of determining the axis of coordinates is the same:
\begin{equation}
    \begin{aligned}
        ^{w}\mathbf{i}_{S,x}&=^{w}\mathbf{i}_{0,2}\\
        ^{w}\mathbf{i}_{S,y}&=^{w}\mathbf{i}_{0,2}\times ^{w}\mathbf{i}_{0,3}\\
        ^{w}\mathbf{i}_{s,z}&=^w\mathbf{i}_{S,x}\times^w\mathbf{i}_{S,y}
    \end{aligned}
\end{equation}

Where $\mathbf{i}_{0,2}, \mathbf{i}_{0,3}$ are unit vectors along robot fingers that correspond to human middle and ring fingers and the index corresponds to the joint numbers defined by MANO. We build the following two coordinate transformation matrix to construct the mapping:
\begin{equation}
    \begin{aligned}
        ^W_H\mathbf{R}&=
\begin{pmatrix}
 ^W\mathbf{I}_H & \mathbf{key}_{human}\\
 \mathbf{O} & 1
\end{pmatrix}\\
^W_{S}\mathbf{R}&=
\begin{pmatrix}
 {^W\mathbf{I}_S} & {\mathbf{key}_{robot}}\\
    \mathbf{O} & {1}
\end{pmatrix}
    \end{aligned}
\end{equation}

Where $\mathbf{key}_{human},\mathbf{key_{robot}}$ are the positions of human wrist and robot wrist. After obtaining the two coordinate systems, we need to determine the position of the camera in the simulator ($^W\mathbf{cam}_{sim}$) and the position of the camera in the real world ($^H\mathbf{cam}_{Real}$), thus we can ensure we get the same pose of the human hand and robot arms
\begin{equation}
    \begin{aligned}
        ^H\mathbf{cam}_{Real}&=^W_H\mathbf{R}^{-1}\times^W\mathbf{cam}_{Real}\\
        ^S\mathbf{cam}_{sim}&=^H\mathbf{cam}_{Real}\\
        ^W\mathbf{cam}_{sim}&=^W_S\mathbf{R}\times^W_H\mathbf{R}^{-1}\times^W\mathbf{cam}_{Real}
    \end{aligned}
\end{equation}

\section{Details of Experimental Setup}
\label{sec:trainingdetails}

\begin{figure*}[!t] 
    \centering
    \includegraphics[width=\textwidth]{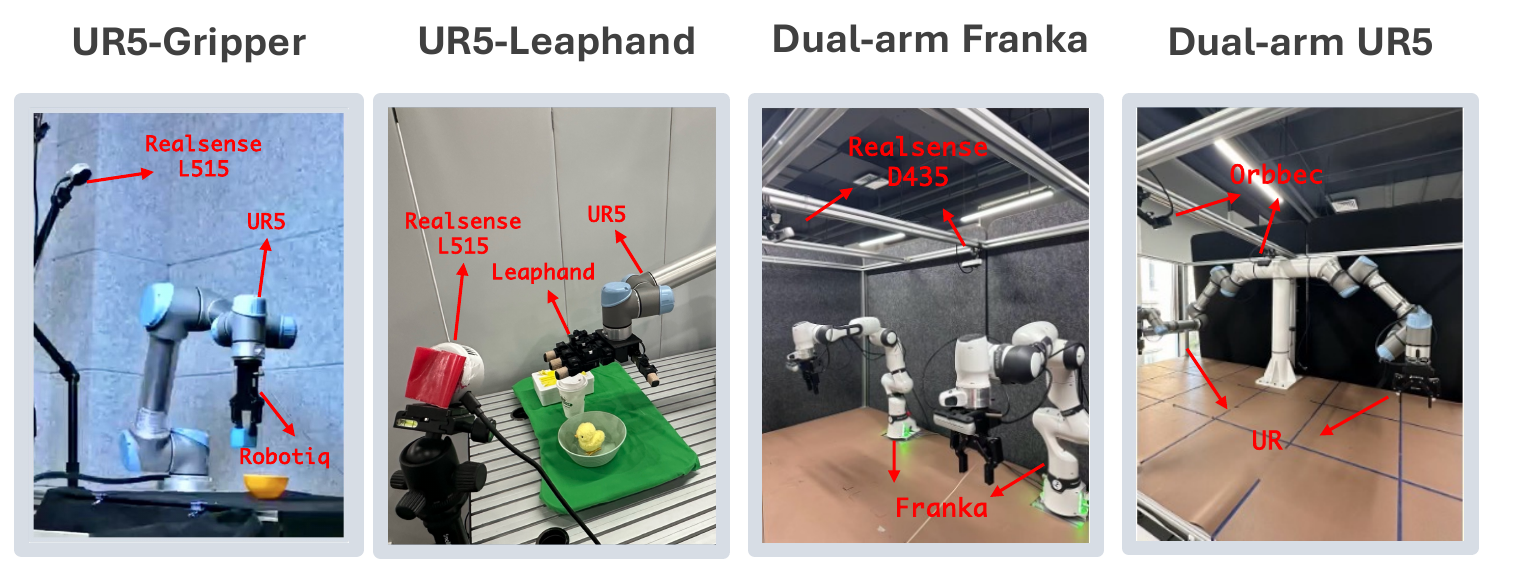}
    \caption{\textbf{Real-world Robot Setup.} Illustration of different real-world experimental setup.} 
    \label{fig:real-world-setup}
\end{figure*}
\begin{table*}[htbp]
    \centering
        \caption{\textbf{Robomimic Experiment result.} We report the success rate (\%) over IL-based tasks for MAE and R3M Robomimic.}
    \begin{adjustbox}{max width=\textwidth}
    \begin{tabular}{c|cccc||c}
    \toprule
     & MoveCan & Square & Lift & Average & PushT \\ \midrule
    MAE & 54 & 25.5 & 94.5 & 58 & 59.2 \\
    MAE+CutMix1 & 72.0 \textcolor{blue}{(+18.0\%)} & 30.0 \textcolor{blue}{(+4.5\%)} & 95.0 \textcolor{blue}{(+0.5\%)} & 65.7 \textcolor{blue}{(+7.7\%)} & 37.5 \textcolor{red}{(-21.7\%)} \\
    MAE+CutMix2 & 58.0 \textcolor{blue}{(+4.0\%)} & 36.0 \textcolor{blue}{(+10.5\%)} & 90.0 \textcolor{red}{(-4.5\%)} & 61.3 \textcolor{blue}{(+3.3\%)} & 40.0 \textcolor{red}{(-19.2\%)} \\
    MAE+CutMix3 & 78.0 \textcolor{blue}{(+24.0\%)} & 32.0 \textcolor{blue}{(+9.3\%)} & 92.0 \textcolor{red}{(-2.5\%)} & 67.3 \textcolor{blue}{(+2.7\%)} & 42.0 \textcolor{red}{(-17.2\%)} \\
    MAE+H2R & 79.5 \textcolor{blue}{(+25.5\%)} & 29.5 \textcolor{blue}{(+4.0\%)} & 95.5 \textcolor{blue}{(+1.0\%)} & 68.2 \textcolor{blue}{(+10.2\%)} & 64.5 \textcolor{blue}{(+5.3\%)} \\ \midrule
    R3M & 59.5 & 20.5 & 85 & 55 & 15 \\
    R3M+CutMix1 & 69.5 \textcolor{blue}{(+10.0\%)} & 30.0 \textcolor{blue}{(+9.5\%)} & 91.0 \textcolor{blue}{(+6.0\%)} & 63.5 \textcolor{blue}{(+8.5\%)} & 19.0 \textcolor{blue}{(+4.0\%)} \\
    R3M+CutMix2 & 66.0 \textcolor{blue}{(+6.5\%)} & 26.0 \textcolor{blue}{(+5.5\%)} & 83.0 \textcolor{red}{(-2.0\%)} & 58.3 \textcolor{blue}{(+3.3\%)} & 17.0 \textcolor{blue}{(+2.0\%)} \\
    R3M+CutMix3 & 68.0 \textcolor{blue}{(+8.5\%)} & 26.0 \textcolor{blue}{(+5.5\%)} & 84.0 \textcolor{red}{(-1.0\%)} & 59.3 \textcolor{blue}{(+4.3\%)} & 14.0 \textcolor{red}{(-1.0\%)} \\
    R3M+H2R & 61.5 \textcolor{blue}{(+2.0\%)} & 37.5 \textcolor{blue}{(+17.0\%)} & 85.0 \textcolor{black}{(0.0\%)} & 61.3 \textcolor{blue}{(+6.3\%)} & 22.0 \textcolor{blue}{(+7.0\%)} \\ \midrule
    \end{tabular}
    \end{adjustbox}
    \label{table:il_ablation}
\end{table*}
\subsection{Additional Evaluation Details.}
Figure ~\ref{fig:simulator} provides visualizations of the different simulation benchmarks, and Figure ~\ref{fig:real-world-setup} illustrates the real-world experimental setup.

\begin{figure*}[htbp]
    \centering
    \begin{minipage}[b]{0.9\textwidth}
        \centering
        \includegraphics[width=\textwidth]{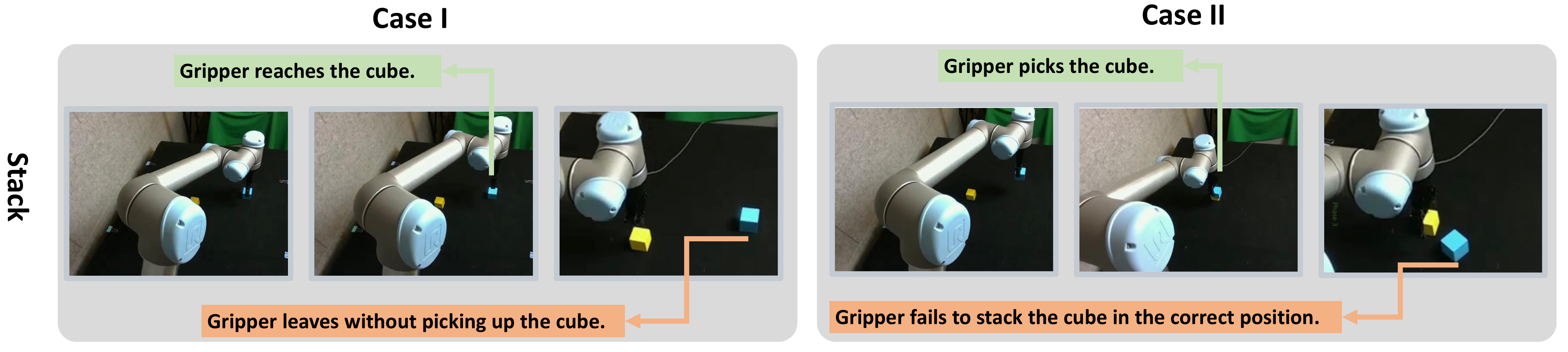}
        \textbf{Stack}
        \label{fig:subfig1}
    \end{minipage}
    \begin{minipage}[b]{0.9\textwidth}
        \centering
        \includegraphics[width=\textwidth]{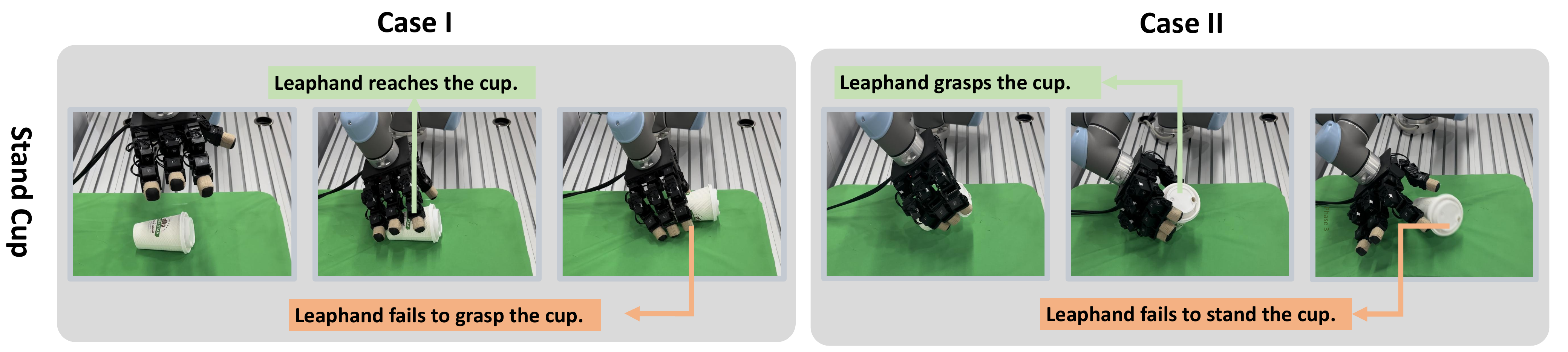}
        \textbf{Stand Cup}
        \label{fig:subfig2}
    \end{minipage}
    \caption{\textbf{Failure case visualizations: Stack and Stand Cup.} We visualize real-world manipulation executions for two downstream tasks: \textit{Stack} (top) and \textit{Stand Cup} (bottom). These images provide qualitative insights into the performance and failure modes of the policy in real deployment, highlighting challenges such as object misalignment, perception noise, and grasp precision.}
    \label{fig:fail}
\end{figure*}
\section{Failure Case Analysis}
\label{sec:failcase}

\begin{table*}[htbp]
\centering
\caption{\textbf{Task-specific sub-goal evaluation.} To gain fine-grained insights into policy performance, we design a manual rubric covering key sub-goals for each manipulation task. Each cell reports the number of successful vs. unsuccessful attempts (\texttt{Y/N}) over 20 evaluation trials. 
Results show that models enhanced with H2R consistently accomplish more sub-goals across tasks compared to their baseline counterparts, demonstrating improved robustness in real-world execution. Bold numbers indicate better performance between paired models.}
\begin{adjustbox}{max width=\textwidth}
\begin{tabular}{c|c|cc|cc}
\toprule
Task & Sub-goal & MAE(Y/N) & MAE+H2R(Y/N) & R3M(Y/N) & R3M+H2R(Y/N) \\ \midrule
\multirow{2}{*}{Gripper-PickCube} 
& Overall success? & 9/11 & \textbf{13}/7 & 8/12 & \textbf{10}/10\\
& Pick up the cube? & 14/6 & \textbf{15}/5 & 11/9 & \textbf{13}/7 \\
 \midrule
\multirow{2}{*}{Gripper-Stack}
& Overall success? &  10/10 & \textbf{11}/9 & 11/9 & \textbf{14}/6 \\ 
& Pick up the cube? &  13/7 & \textbf{16}/4 & 13/7 & \textbf{17}/3 \\ 
 \midrule
\multirow{3}{*}{Gripper-CloseBox}
& Overall success? &  \textbf{11}/9 & 10/10 & 9/11 & \textbf{13}/7 \\ 
& Place the cube in the bow? &  12/8 & \textbf{14}/6 & 12/8 & \textbf{15}/5 \\ 
& Pick up the cube? &  \textbf{14}/6 & \textbf{14}/6 & 12/8 & \textbf{15}/5 \\ 
 \midrule
\multirow{2}{*}{Leaphand-GraspChicken}
& Overall success? &  8/12 & \textbf{11}/9 & 2/18 & \textbf{7}/13 \\ 
& Pick up the chicken? &  13/7 & \textbf{14}/6 & 3/17 & \textbf{10}/10 \\ 
 \midrule
\multirow{2}{*}{Leaphand-StandCup}
& Overall success? &  7/13 & \textbf{12}/8 & 4/16 & \textbf{10}/10 \\ 
& Pick up the cup? &  12/8 & \textbf{18}/2 & 12/8 & \textbf{15}/5 \\ 
 \midrule
\multirow{2}{*}{Leaphand-OpenBox}
& Overall success? &  9/11 & \textbf{13}/7 & 8/12 & \textbf{9}/11 \\ 
& Identify contact location? &  14/6 & \textbf{16}/4 & \textbf{10}/10 & \textbf{10}/10 \\ 
 \bottomrule
\end{tabular}
\end{adjustbox}
\label{table:sub-goal}
\end{table*}
To better understand the limitations of our policy and the challenges encountered in real-world deployments, we present a qualitative analysis of failure cases from two representative tasks: a gripper-based task (\textit{Gripper-Stack}) and a dexterous manipulation task (\textit{Gripper-StandCup}). Figure~\ref{fig:fail} illustrates typical failure modes observed during execution.

In the \textbf{Gripper-Stack} task, we identify two major failure scenarios:

\textbf{Case I: Grasp Failure Unnoticed.} The robot arm fails to successfully grasp the blue cube. However, the policy proceeds as if the object had been grasped, moving toward the yellow cube and attempting to perform the stacking operation. This leads to a complete task failure.

\textbf{Case II: Misaligned Placement.} The robot successfully grasps the blue cube but fails to align it correctly on top of the yellow cube during the stacking phase, resulting in an unstable or failed placement.

In the \textbf{Stand Cup} task, similar issues emerge due to perception and control limitations:

\textbf{Case I: Grasp Position Error.} The Leaphand end-effector attempts to grasp the cup but fails to target the correct contact region. As a result, the cup slips out of the grasp during lifting, preventing task completion.

\textbf{Case II: Insufficient Lifting Trajectory.} Even when the grasp is successful, the lifting motion lacks sufficient amplitude or stability to fully stand the cup upright. The cup either tips over or fails to stand securely.

To enable fine-grained evaluation of policy performance and gain deeper insights into failure cases, we designed a task-specific evaluation rubric. Table~\ref{table:sub-goal} displays our rubric that the evaluator filled out when rolling out different policies. Take the DP policy as an example, the results in Table~\ref{table:sub-goal} demonstrate that H2R-augmented visual representation models not only improve overall success rates in real-world tasks, but also allow to accomplish more than half of the task consistently.




\section{Additional Ablation Study in Simulation Experiment}
\label{sec:simablation}
In addition to pre-training on the H2R data and raw data, we also applied a simple CutMix baseline to demonstrate the effectiveness of using the robotic arm to cover the human hand, which overlays a fixed set of specific images of robotic arms with grippers onto the original images, ensuring that the overlaid images cover the human hands as much as possible, without exceeding the detected bounding box. Our H2R is different from such baseline by employing robot hand construction to better match the pose of the hand and arm in the images. Based on the type of robotic arm used in CutMix, we categorize the augmented set into three types: CutMix1 represents the UR5 robotic arm, CutMix2 refers to the Franka robotic arm, and CutMix3 combines both the UR5 and Franka robotic arms.

From Table \ref{table:il_ablation}, we observe that the encoder trained on H2R processed data shows consistent improvements across various tasks compared to the encoder trained on the original data, with the average success rate on all tasks ranging from 0.9\% to 10.2\%. Especially for the more challenging MoveCan task, it can improve the success rate by 25.5\%. Additionally, while encoders trained on the relatively simple CutMix data show improvement on tasks in Robomimic, their performance in the PushT task remains slightly worse than the encoders trained on original data. These results demonstrate the effectiveness of using the robotic arm to cover the human hand in video data, as well as the effectiveness of \ours in imitation learning.


\end{document}